\documentclass[journal,11pt]{IEEEtran} 

\usepackage{graphicx}
\usepackage{sidecap}
\usepackage{xcolor}
\usepackage{acronym}
\usepackage[nonumberlist,nogroupskip]{glossaries}
\usepackage{glossary-mcols}
\usepackage{comment}
\usepackage{wrapfig}
\usepackage{setspace}
\usepackage[numbers, square, comma, sort&compress]{natbib}
\usepackage[font=small]{caption}
\usepackage{mdframed}
\usepackage{subfig}
\usepackage{amssymb}
\DeclareCaptionType[fileext=box]{infobox}[Box]
\usepackage[utf8]{inputenc}
\usepackage{subfig}

\newcommand{\refeq}[1]{{Equation~(\ref{#1})}}

\newcommand{\reffig}[1]{{\color{blue!70}(Fig.~\ref{#1})}}

\newcommand{\Dp}[2][]{\frac{\partial #1}{\partial #2}}

\newcommand{\pd}[2]{\frac{\partial #1}{\partial #2}}

\begin{document}
\title{Brain-Inspired Learning on Neuromorphic Substrates}
\acrodef{IR}[IR]{Intrinsic Rewards and Motivation}
\acrodef{PPO}[PPO]{Proximal Policy Optimization}
\acrodef{RL}[RL]{Reinforcement Learning}
\acrodef{AC}[AC]{Arrenhius \& Current}
\acrodef{AD}[AD]{Automatic Differentiation}
\acrodef{AER}[AER]{Address Event Representation}
\acrodef{AEX}[AEX]{AER EXtension board}
\acrodef{AMDA}[AMDA]{``AER Motherboard with D/A converters''}
\acrodef{ANN}[ANN]{Artificial Neural Network}
\acrodef{API}[API]{Application Programming Interface}
\acrodef{BP}[BP]{Back-Propagation}
\acrodef{BPTT}[BPTT]{Back-Propagation-Through-Time}
\acrodef{BM}[BM]{Boltzmann Machine}
\acrodef{CAVIAR}[CAVIAR]{Convolution AER Vision Architecture for Real-Time}
\acrodef{CCN}[CCN]{Cooperative and Competitive Network}
\acrodef{CD}[CD]{Contrastive Divergence}
\acrodef{CG}[CG]{Computational Graph}
\acrodef{CMOS}[CMOS]{Complementary Metal--Oxide--Semiconductor}
\acrodef{CNN}[CNN]{Convolutional Neural Network}
\acrodef{COTS}[COTS]{Commercial Off-The-Shelf}
\acrodef{CPU}[CPU]{Central Processing Unit}
\acrodef{CV}[CV]{Coefficient of Variation}
\acrodef{CTC}[CTC]{connectionist temporal classification}
\acrodef{DAC}[DAC]{Digital--to--Analog}
\acrodef{DBN}[DBN]{Deep Belief Network}
\acrodef{DCLL}[DECOLLE]{Deep Continuous Local Learning}
\acrodef{DFA}[DFA]{Deterministic Finite Automaton}
\acrodef{DFA}[DFA]{Deterministic Finite Automaton}
\acrodef{divmod3}[DIVMOD3]{divisibility of a number by 3}
\acrodef{DPE}[DPE]{Dynamic Parameter Estimation}
\acrodef{DPI}[DPI]{Differential-Pair Integrator}
\acrodef{DSP}[DSP]{Digital Signal Processor}
\acrodef{DVS}[DVS]{Dynamic Vision Sensor}
\acrodef{EDVAC}[EDVAC]{Electronic Discrete Variable Automatic Computer}
\acrodef{EIF}[EI\&F]{Exponential Integrate \& Fire}
\acrodef{EIN}[EIN]{Excitatory--Inhibitory Network}
\acrodef{EPSC}[EPSC]{Excitatory Post-Synaptic Current}
\acrodef{EPSP}[EPSP]{Excitatory Post--Synaptic Potential}
\acrodef{eRBP}[eRBP]{Event-Driven Random Back-Propagation}
\acrodef{FPGA}[FPGA]{Field Programmable Gate Array}
\acrodef{FSM}[FSM]{Finite State Machine}
\acrodef{GPU}[GPU]{Graphical Processing Unit}
\acrodef{HAL}[HAL]{Hardware Abstraction Layer}
\acrodef{HH}[H\&H]{Hodgkin \& Huxley}
\acrodef{HMM}[HMM]{Hidden Markov Model}
\acrodef{HW}[HW]{Hardware}
\acrodef{hWTA}[hWTA]{Hard Winner--Take--All}
\acrodef{IF2DWTA}[IF2DWTA]{Integrate \& Fire 2--Dimensional WTA}
\acrodef{IF}[I\&F]{Integrate \& Fire}
\acrodef{IFSLWTA}[IFSLWTA]{Integrate \& Fire Stop Learning WTA}
\acrodef{INCF}[INCF]{International Neuroinformatics Coordinating Facility}
\acrodef{INRC}[INRC]{Intel Neuromorphic Research Community}
\acrodef{INI}[INI]{Institute of Neuroinformatics}
\acrodef{IO}[IO]{Input-Output}
\acrodef{IoT}[IoT]{internet of things}
\acrodef{IPSC}[IPSC]{Inhibitory Post-Synaptic Current}
\acrodef{ISI}[ISI]{Inter--Spike Interval}
\acrodef{JFLAP}[JFLAP]{Java - Formal Languages and Automata Package}
\acrodef{LIF}[LIF]{Linear Integrate and Fire}
\acrodef{LSM}[LSM]{Liquid State Machine}
\acrodef{LTD}[LTD]{Long-Term Depression}
\acrodef{LTI}[LTI]{Linear Time-Invariant}
\acrodef{LTP}[LTP]{Long-Term Potentiation}
\acrodef{LTU}[LTU]{Linear Threshold Unit}
\acrodef{LSTM}[LSTM]{long short-term memory}
\acrodef{MCMC}{Markov Chain Monte Carlo}
\acrodef{MSE}{Mean-Squared Error}
\acrodef{NHML}[NHML]{Neuromorphic Hardware Mark-up Language}
\acrodef{NMDA}[NMDA]{NMDA}
\acrodef{NME}[NE]{Neuromorphic Engineering}
\acrodef{PCB}[PCB]{Printed Circuit Board}
\acrodef{PRC}[PRC]{Phase Response Curve}
\acrodef{PSC}[PSC]{Post-Synaptic Current}
\acrodef{PSP}[PSP]{Post--Synaptic Potential}
\acrodef{RI}[KL]{Kullback-Leibler}
\acrodef{RRAM}[RRAM]{Resistive Random-Access Memory}
\acrodef{RBM}[RBM]{Restricted Boltzmann Machine}
\acrodef{RTRL}[RTRL]{Real-Time Recurrent Learning}
\acrodef{ROC}[ROC]{Receiver Operator Characteristic}
\acrodef{RSA}[RSA]{Representational Similarity Analysis}
\acrodef{RDA}[RDA]{Representational Dissimilarity Analysis}
\acrodef{RDM}[RDA]{Representational Dissimilarity Matrix}
\acrodef{RNN}[RNN]{Recurrent Neural Network}
\acrodef{SAC}[SAC]{Selective Attention Chip}
\acrodef{SCD}[SCD]{Spike-Based Contrastive Divergence}
\acrodef{SCX}[SCX]{Silicon CorteX}
\acrodef{SG}[SG]{Surrogate Gradient}
\acrodef{SGD}[SGD]{Surrogate Gradient Descent}
\acrodef{SRM}[SRM]{Spike Response Model}
\acrodef{SNN}[SNN]{Spiking Neural Network}
\acrodef{STDP}[STDP]{Spike Time Dependent Plasticity}
\acrodef{SW}[SW]{Software}
\acrodef{sWTA}[SWTA]{Soft Winner--Take--All}
\acrodef{TPU}[TPU]{Tensorflow Processing Unit}
\acrodef{VHDL}[VHDL]{VHSIC Hardware Description Language}
\acrodef{VLSI}[VLSI]{Very  Large  Scale  Integration}
\acrodef{WTA}[WTA]{Winner--Take--All}
\acrodef{XML}[XML]{eXtensible Mark-up Language}

\author{Friedemann Zenke$^\dagger$,  Emre O. Neftci$^\dagger$\\
        {\small $^\dagger$ All authors contributed equally}}%
\markboth{}%
{}
\maketitle

\begin{abstract}
Neuromorphic hardware strives to emulate brain-like neural networks and thus holds the promise for scalable, low-power information processing on temporal data streams.  Yet, to solve real-world problems, these networks need to be trained.  However, training on neuromorphic substrates creates significant challenges due to the offline character and the required non-local computations of gradient-based learning algorithms.  
This article provides a mathematical framework for the design of practical online learning algorithms for neuromorphic substrates.   Specifically, we show a direct connection between \ac{RTRL}, an online algorithm for computing gradients in conventional \acp{RNN}, and biologically plausible learning rules for training \acp{SNN}.   
Further, we motivate a sparse approximation based on block-diagonal Jacobians, which reduces the algorithm's computational complexity, diminishes the non-local information requirements, and empirically leads to good learning performance, thereby improving its applicability to neuromorphic substrates.  
In summary, our framework bridges the gap between synaptic plasticity and gradient-based approaches from deep learning and lays the foundations for powerful information processing on future neuromorphic hardware systems.
\end{abstract}
\acresetall
\IEEEpeerreviewmaketitle

\section{Introduction}
Our brains simultaneously process various streams of temporal information allowing us to solve challenging real-world problems that ensure our survival.
Importantly, brains do this with an aptitude that dwarfs existing computer technologies while only consuming a meek 25W of power.

Neuromorphic engineering has taken on the challenge of approaching such efficiency by building scalable, low-power systems that mirror the brain's essential architectural features \cite{Mead89_analvlsi,Thakur_etal18_largneur,Davies19_bencprog}. 
Decades of research helped overcome major engineering challenges toward this goal, resulting in an increasing number of neuromorphic substrates available today  \cite{Merolla_etal14_millspik, Davies_etal18_loihneur, Indiveri_etal11_neursili} and allowing the efficient emulation of brain-inspired neural networks. 
One key challenge that remains and prevents the widespread application of neuromorphic systems is the lack of practical algorithms that run on such hardware and equip it with complex functionality.

Deep learning provides algorithmic blueprints to organize large neural networks into suitable function approximators that flexibly solve diverse real-world problems \cite{Goodfellow_etal16_deeplear}.
To achieve this feat, deep learning optimizes loss functions with gradient descent, which can be computed efficiently with the \ac{BP} algorithm.
This efficiency, however, rests on the von Neumann computer architecture. 
In contrast, gradient \ac{BP} is difficult to implement on non-von Neumann neuromorphic substrates. 
These difficulties mainly arise from limitations in their ability to 
communicate neural activities and weight values between different network elements, similar to the architectural constraints of biological neural networks \citep{lillicrap_backpropagation_2020}.
For instance, synaptic plasticity implemented at a biological synapse may have access to the activity of the two
neurons it connects, but not to the activity of other neurons that it is not physically connected to. 
This notion is often expressed by saying that plasticity in the brain is \emph{local}. 
Local learning rules have been extensively studied in computational
neuroscience, typically based on experimental data.
However, these rules often lack the normative foundations of \ac{BP} and hence the ability to instantiate complex functional neural networks.
However, top-down driven synaptic plasticity rules can also be derived from gradient descent both in the case of single biologically inspired spiking neurons
\cite{gutig_tempotron:_2006, Bohte_etal00_spikback, pfister_optimal_2006},
thereby establishing a link to the Perceptron \cite{Rosenblatt58_percprob}, and
in more complex multi-layer networks \cite{rezende_stochastic_2014, gardner_learning_2015, Shrestha_Orchard18_slayspik, Lee_etal16_traideep, zenke_superspike:_2018, bellec_long_2018, kaiser_synaptic_2020, neftci_surrogate_2019, wozniak_deep_2020}.
Remarkably, many normative approaches do result in learning rules largely consistent with models of cortical neurons and synapses 
\cite{Sacramento_etal18_dendcort, brea_matching_2013,
fremaux_neuromodulated_2016, kusmierz_learning_2017, 
guerguiev_towards_2017, zenke_superspike:_2018,  
bellec_solution_2020}. 

\begin{figure}
    \centering
    \includegraphics{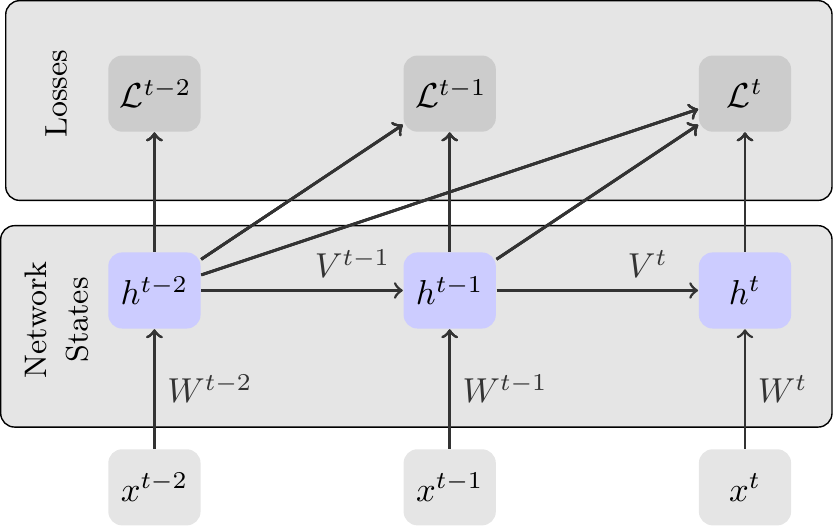}
	\caption{Computational graph of an \acf{RNN}. The bottom row represents the inputs $x^t$.
    The middle row represents the network states of a network with parameters $\theta=\{W,V\}$. 
	Here, arrows indicate operations on the nodes, and the arrow labels indicate
	the parameters involved in the computations. 
	Note that to distinguish between direct and indirect influence, the
	parameters have superscripts with time indices, even though these parameters are tied across time. 
	The top row shows the computations of the output loss in each time step.}
    \label{fig:loss_and_dynamics}
\end{figure}

In the present article, we discuss the remarkable commonalities across machine learning and computational neuroscience learning algorithms from the standpoint of neuromorphic engineering.
To that end, we rely on the conceptual framework provided by deep learning allowing us to focus on three distinct aspects of building functional artificial neural networks: architecture, learning rules, and loss functions (Fig.~\ref{fig:loss_and_dynamics}; \cite{richards_deep_2019}).
Within this framework, we focus on online learning rules (Sections~\ref{sec:learning} \& \ref{sec:bio-inspired-learning}) and loss functions (Section~\ref{sec:losses}) due to their specific relevance for neuromorphic engineering.
This relevance largely derives from the use of non-von Neumann architectures in this field, which we introduce next. 

\subsection{Von Neumann computers and biological brains}

Conceptually, computation requires memory, communication, and arithmetic.  
Computation is a physical process in which these three components come together in the same place and at the same time. 
Doing so requires space, time, and energy. 
And the amount of resources used determines the overall cost of computation,
and varies depending on both the computation itself and the architecture on
which it runs. 

The von Neumann architecture, which underlies virtually all human-made
computers, posits a physical separation between processing units for arithmetic, program flow control, and memory. 
This separation induces a communication channel between the two. 
A significant design limitation of this architecture is that when the amount of data becomes too large, communication between the processing and the memory unit becomes a bottleneck. 
Thus, for memory-intensive computations, this so-called \emph{von Neumann
Bottleneck} impacts both latency and power efficiency.
The impact is especially noticeable in deep learning applications that require
large amounts of data to first propagate through immense deep neural network models. 
Then gradient information propagates back through the same nodes to compute
gradients for learning. 
Since both the data and the model parameters reside in memory while the
operations that act on them take place in the processing units, the
communication needs become substantial.
Although modern computer architectures use diverse strategies to mitigate this
bottleneck, e.g., caching, branch prediction, and parallelism, all of these
measures come at the expense of higher energy cost and larger on-chip space
requirements.
Hence, they eventually run into the same bottleneck. 
With the looming end of Moore's law, the communication bandwidth across processing and memory units is reaching a plateau.
While the continued development of deep learning hinges in large part on ever faster and larger-scale hardware, 
some emerging neuromorphic developments seek to provide alternatives to the von Neumann architecture \cite{Neftci18_datapowe}.

While von Neumann architectures \emph{simulate} deep neural networks, neuromorphic solutions attempt to \emph{emulate} them on a physical substrate, inspired by the brain.
In the brain, neurons integrate inputs in an analog manner, apply nonlinear transformations to them, and communicate asynchronously through digital spikes or action potentials. 
Spikes are binary events in continuous time that evoke electric currents in receiving neurons whose strength is determined by their connecting synapses.
Importantly, these connections are plastic and can change dynamically in an experience-dependent manner to implement learning and memory. 
Together, neurons and their intricate synaptic connectivity achieve neural processing and long-term storage in a completely parallel and time-continuous way.   
Since memory and arithmetic are physically co-localized and distributed within the same physical network, brains are fundamentally non-von Neumann architectures. 
Sparse, event-based communications and physically co-localizing memory and computation are two defining principles of brain-inspired computing and neuromorphic engineering \cite{Neftci18_datapowe, sterling_principles_2017}. 
Neuromorphic engineers recognized the benefits of this approach and developed large networks of VLSI neurons and synapses communicating asynchronously through \ac{AER} \cite{Lazzaro_etal93_siliaudi, Deiss_etal98_pulscomm}.
Recent digital hardware, such as Google's TPU \cite{Jouppi_etal17_in-dperf}, Graphcore's Intelligence Processing Unit (IPU), and Cerebras' wafer-scale CS-1, have also embraced some form of memory--computation co-localization to improve the energy and performance metrics of scientific computing and machine learning workloads.
The development of these brain-inspired architectures is in large part motivated by the need to process the ever-increasing deluge of data generated by the pervasive sensors in everyday life.

\subsection{Processing and learning from temporal data streams}

A large fraction of sensor data are temporal and require real-time processing, thus warranting dedicated architectures.
For example, autonomous vehicles need continuous monitoring and processing of time series data from their sensors to navigate safely. Similarly, \ac{IoT} devices have to continuously monitor their environment to respond to speech commands, detect anomalies online from biosensor data, (\emph{e.g.}, a pacemaker implant), and to learn continually \citep{kirkpatrick_overcoming_2017,Zenke_etal17_imprmult, parisi_continual_2018}. 
While data streams induce specific challenges, most of which we will discuss in Section~\ref{sec:losses}, we first focus on the essential
fact that most real-world data, whether streaming or not, is also temporal data that needs to be processed in real-time. 
In the following, we briefly review \acp{RNN} as the de-facto standard for processing temporal data before focusing specifically on neuromorphic implementations thereof and how we can accomplish gradient-based optimization in real-time. 

The brain is an inherently time-dependent dynamical system \cite{drion_neuronal_2015,izhikevich_dynamical_2007} that relies on biophysical processes, recurrence, and feedback of its physical substrate for computation \citep{bhalla_molecular_2014,sterling_principles_2017}. 
These dynamics are different from the majority of deep neural networks, which are often strictly feedforward, and lack the fine temporal dynamics of brains. 
From a technological point of view, emulating neural dynamics on a physical substrate has the advantage of operating much more efficiently compared to simulations \cite{Mead90_neurelec}.
However, this comes with a key challenge in which ``time represents itself.'' 
This implies that all computational processes occur at the timescales of the physical system. 
In VLSI technologies, this is achieved by operating CMOS transistors in their subthreshold regime, such that currents and consequently the time constants are matched to those of the brain \cite{Livi_Indiveri09_currcond, Mead90_neurelec}. 
As such, these systems are both online and streaming.
Other technologies run in accelerated time \citep{schemmel_accelerated_2017}, which can create a mismatch of time scales in certain real-world applications.

\acp{RNN} have proven highly effective for sequential processing such as keyword spotting, object recognition, or time series forecasting \cite{Goodfellow_etal16_deeplear,Graves12_supesequ}.  
Neuromorphic processors generally implement \acp{SNN} \cite{Indiveri_etal11_neursili}, which can be viewed as a special class of \acp{RNN} inspired by biology.
\acp{SNN} are particularly suited for energy-efficient processing thanks to their rich local dynamics but spatiotemporally sparse communication via spikes (events).
This is in contrast to most \acp{RNN} used in machine learning which rely on dense and analog valued communications.
Moreover, biological neurons presumably carry specific inductive biases in their internal dynamics that are potentially advantageous for real-world information processing.

Like all neural networks, \acp{SNN} can be trained to find suitable connection weights.
Because \acp{SNN} are \acp{RNN}, they can be trained with similar gradient-based methods that only need to be modified slightly to accommodate the binary activation functions underlying action potentials \citep{neftci_surrogate_2019}.
Gradient descent on a loss function thereby automatically adjusts neuron and synapse parameters in the hidden layers of the network to reduce a scalar training loss.
In the following, we review two common algorithms underlying gradient computation in \acp{RNN}, and highlight their limitations for gradient-based learning in neuromorphic substrates, before discussing a range of possible remedies.

\section{Gradient-based Learning in Recurrent Neural Networks}
\label{sec:learning}
Training \acp{RNN} requires computing objective function gradients with respect to the network parameters.
To gain a better understanding of why specific challenges arise when computing gradients for \acp{RNN}, we consider a simple \ac{RNN} 
\[
h^{t}=f_{\theta}(h^{t-1},x^{t})
\]
with the network state $h^{t}\in\mathcal{\mathbb{R}}^{k}$, the input $x^{t}\in\mathbb{R}^{n}$ and the parameters $\theta\in\mathbb{R}^{p}$, where $p$ is the number of parameters. 
We further define the output $y^{t}=g_{\theta}(h^{t})$, and the associated target $y^{t*}$ and loss function $\mathcal{L}=\sum_{t}\mathcal{L}^{t}(y^{t},y^{t*})$. 
Training an \ac{RNN} requires computing the gradients of $\mathcal{L}$ with respect to all the parameters $\theta$.
Following the steps of Marschall \emph{et al.} \citep{Marschall_etal19_uniffram}, we define $\theta^t$ as the application of the parameter at time $t$ to distinguish between their direct and indirect influence, and note that all parameters $\theta^t$ are tied across all timesteps $t$. 
The gradient is given by:
\begin{equation}
\pd{\mathcal{L}}{\theta} = \sum_t \pd{\mathcal{L}^{t}}{\theta} = 
\sum_t \pd{\mathcal{L}^t}{h^t}\pd{h^t}{\theta} =
\sum_t \pd{\mathcal{L}^t}{h^t} \sum_{s \le t} \pd{h^t}{\theta^s}. 
\label{eq:grad}
\end{equation}
Two different temporal summations appear in this expression. 
To be able to use Equation~\eqref{eq:grad} for online learning, we have to first evaluate the sum over $s$ which underlies $\pd{h^t}{\theta}$.
Fortunately, it is possible to compute $\pd{h^t}{\theta}$ with the help of a simple recursion relationship. 
To write this relationship compactly, we define the influence $G^{t}:=\pd{h^{t}}{\theta} \in \mathbb{R}^{k\times p}$, the immediate influence $F^{t}:=\frac{\partial h^{t}}{\partial\theta^t} \in \mathbb{R}^{k\times p}$,
and the \ac{RNN}'s dynamics $H^{t}:=\frac{\partial h^{t}}{\partial h^{t-1}}\in \mathbb{R}^{k\times k}$.
By further assuming $G^{t=0}=0$, $G^t$ is given as the recursive product of Jacobians:
\begin{equation}
G^{t} = H^{t} G^{t-1}+ F^{t}, 
\label{eq:rtrl}
\end{equation}
which can be computed going forward in time.
For a detailed derivation of the above expressions, see \citep{Marschall_etal19_uniffram}.
By inserting $G^t$ back into \refeq{eq:grad}, and assuming a small learning rate, the summation over $t$ can be implemented as an online learning algorithm.
This causal learning algorithm is called \ac{RTRL} \cite{williams_learning_1989, Marschall_etal19_uniffram}.

However, a more common method to evaluate the gradient is \ac{BPTT}.
\ac{BPTT} takes an acausal approach to evaluate the same Jacobian matrix products: it computes the product of Jacobians starting at the end and working its way backward through time, hence its name.
To make this relationship explicit we define
the credit assignment vector $C^t=\pd{\mathcal{L}}{h^t} \in \mathbb{R}^{k}$ and the instantaneous credit vector $D^t=\pd{\mathcal{L}^t}{h^t} \in \mathbb{R}^{k}$. 
The recursion relation underlying \ac{BPTT} is then given by:
\begin{equation}
  C^t = C^{t+1}H^{t+1} + D^{t}.
\label{eq:bptt}
\end{equation}
This expression needs to be computed in an acausal
manner \citep{Marschall_etal19_uniffram}, which precludes its use as an online learning algorithm.
Although, for a given sequence, the gradients resulting from \ac{BPTT} and \ac{RTRL} are
the same, \ac{BPTT} remains the gold standard for training \acp{RNN} on von Neumann hardware.
The reason is that the different implementations have specific advantages. 
To understand the origin of these differences, we now analyze the computational costs of \ac{BPTT} and \ac{RTRL}.

\subsection{Cost analysis of \ac{BPTT}}

The first term in \refeq{eq:bptt} is the $k$-vector derivative of a scalar loss function, and is thus a row vector.
The factorization afforded by the chain rule means that all products are Vector Jacobian products of dimensions $k$ and $k\times k$, respectively, and can be computed in $k^2$ evaluations. 
However, to evaluate the gradient in reverse-mode, it is necessary to record all evaluations in the forward phase. 
Thus, the full activation history needs to be stored in memory, and in the case of \acp{RNN}, memory requirement scales as $O(k T)$. 
The time complexity of each timestep is dominated by the number of scalar multiplications operations underlying the product of the Jacobians which is $O(k^2 T)$ (cf.\ Fig.~\ref{fig:rnn_unrolled}).
The dependence on $T$ restricts \ac{BPTT} to temporal inputs of limited duration and to substrates that offer enough memory to store the activation history. Both requirements are major shortcomings when we want to process continuous streaming data on non-von Neumann architectures on which locally accessible memory is limited.
To reduce the complexity of \ac{BPTT}, gradient propagation is generally truncated at a number of steps smaller than $T$.
This temporal restriction improves complexity, but severely restricts learning performance on tasks that require long time horizons.
Truncation is even suspected to render \acp{RNN} trained with \ac{BPTT} to what is effectively a feed-forward network \cite{Miller_Hardt19_stabrecu}. 

\subsection{Cost Analysis of \ac{RTRL}}
The computational cost of \ac{RTRL} is determined by evaluating $G^t$ as a product of Jacobians of the shape $\mathbb{R}^{k\times k}$ and $\mathbb{R}^{k\times p}$ which requires $k^2 p$ scalar multiplications. 
Since $p$ is the number of parameters, the cost is sizable.
\ac{RTRL} requires $O(k p)$ memory to store $G^t$, a $\mathbb{R}^{k\times p}$ matrix.
For instance, in a fully connected network, we have $k\left(n+k\right)$ parameters for the feedforward and recurrent connections. 
Assuming that $n\sim k$, the overall memory complexity scales cubically with the number of neurons ($O(k^3)$) whereas time scales as $O(k^4)$ per update step. 

Nevertheless, one decisive advantage of \ac{RTRL} is that, although $G^{t}$ is a potentially large matrix, it can be discarded after each update. 
In other words, \ac{RTRL}'s complexity is independent of~$T$, which makes it an interesting contender for training on streaming data on neuromorphic substrates. 
Despite the benefit of online evaluation, \ac{RTRL}'s high computational burden $O(k^4)$, compared to $O(k^2 T)$ for \ac{BPTT}, makes it prohibitive for any practical \ac{RNN} implementation.
\subsection{Reducing the cost of online learning through sparseness}

Fortunately, there are multiple ways to reduce the computational load of \ac{RTRL} while at the same time retaining most of its efficiency.  One way of lowering \ac{RTRL}'s high computational cost is to approximate the computation of the influence matrix $G^{t}$, for instance, by decomposing it in a product of lower-order tensors \citep{tallec_unbiased_2017,cooijmans_variance_2019} or Kronecker factors \citep{mujika_approximating_2018} (see \cite{Marschall_etal19_uniffram} for a comprehensive review). Another way of reducing \ac{RTRL}'s computational cost is to consider sparse approximations of the influence matrix $G^t$ \cite{menick_practical_2020}. 
This situation arises naturally when the network connectivity is either sparse \citep{bellec_deep_2017, frankle_lottery_2018, tanaka_pruning_2020, evci_rigging_2019, liu_finding_2020} or approximately sparse. By approximately, we mean that there are a few strong connections that dominate the temporal dynamics and thus the gradients. 

 Sparsity has additional benefits for neuromorphic substrates because it helps to alleviate two of their inherent limitations. 
 First, it caters to limited on-device storage by requiring less memory for model parameters.
 Second, sparseness can help to overcome communication bottlenecks that result from the need of communicating non-local, but learning-relevant information to where it is needed.  In both \ac{BPTT} and \ac{RTRL}, the update of a single network parameter indirectly depends on all other network parameters, thus rendering learning non-local. 

Due to the relevance of sparseness and local learning rules to neuromorphic substrates, we dedicate the remainder of this article to efficient biologically inspired approximations of \ac{RTRL} and describe how they empower neuromorphic devices to learn from streaming data. 
As we will see, the dynamics of biological neuron models naturally admit a formulation with sparse block Jacobians, which are both sparse and obey locality principles by tying gradient propagation to diagonal blocks that implement the underlying neural and synaptic dynamics. 
But before we can fully appreciate the algorithmic importance of sparse Jacobians, we will briefly review the notion of auto-differentiation, which most modern gradient-based learning algorithms rely on.

\subsection{Auto-differentiation strategies in practice}
\ac{AD} is a type of differentiable programming allowing to compute the gradients across entire computational pipelines \cite{Griewank_Walther08_evalderi,Naumann11_artdiff,Baydin_etal17_autodiff}. 
\ac{AD} has been used extensively in scientific computing before it was applied to neural networks \cite{Bottou_LeCun88_snsimu},
but it has been popularized with recent libraries such as Theano \cite{theano_development_team_theano_2016}, Tensorflow \cite{Abadi_etal15_tenslarg}, and PyTorch \cite{Paszke_etal17_autodiff}.
The key principle underlying \ac{AD} is that numerical computations
are compositions of a finite set of elementary operations for
which derivatives can be defined. 
By combining the derivatives of the operations through the chain rule of derivatives, the gradient of the overall composition is systematically computed.

Just as in Equations \eqref{eq:rtrl} and \eqref{eq:bptt}, gradients in \ac{AD} can be accumulated according to different modes. 
The two main modes are forward and reverse. 
For \acp{RNN}, \emph{forward-mode} accumulation is equivalent to \ac{RTRL}, and \emph{backward-mode} accumulation corresponds to \ac{BPTT}. 
These two modes are two extremes for gradient accumulation.
In principle, it is possible to mix forward and backward within the same algorithm \cite{Revels_etal18_dynaauto,Naumann08_optijaco}.
But for reasons relating to its superior time and space complexity, most machine learning frameworks use pure reverse mode accumulation.
Later in this article, we discuss an example of mixed \ac{AD} modes that is computationally advantageous for neuromorphic hardware.

\section{Applying Gradient-based Learning to Biologically-Inspired Neural Networks} 
\label{sec:bio-inspired-learning}
Biological neural networks are \acp{RNN}, but there are two defining characteristics that set them apart from the generic \ac{RNN} models discussed in the previous section. 
First, biological neurons possess internal dynamics on different timescales due to a plethora of bio-chemical processes that interact with the membrane dynamics \cite{pozzorini_temporal_2013, sterling_principles_2017}. 
Because these dynamics play an important role for the approximations of the Jacobians $H^t$ and $F^t$, which are the basis for the efficient online learning algorithms, we will discuss them in detail in Section~\ref{sec:types_of_recurrence}.
Second, most biological neurons communicate through action potentials, or spikes.
Since spikes are binary neuronal outputs, they render \acp{SNN} non-differentiable, which 
poses a problem for standard gradient-based optimization algorithms. 
Hence, special training methods are required, which we briefly review in the following.

\subsection{Training spiking neural networks}
\label{sec:training_approaches}
Several learning schemes have been developed to overcome the non-differential nature of spiking neurons \cite{abbott_building_2016, pfeiffer_deep_2018, tavanaei_deep_2018, neftci_surrogate_2019}. 
The most commonly used gradient-based \ac{SNN} learning paradigms are network translation \cite{rueckauer_conversion_2017, zambrano_efficient_2017, kim_simple_2019, stockl_classifying_2020},
variational learning with stochastic neuron models \cite{ackley_learning_1985, brea_matching_2013, rezende_stochastic_2014, jang_introduction_2019},
and surrogate gradients in combination with deterministic \ac{LIF} neurons \cite{Lee_etal16_traideep, zenke_superspike:_2018, Shrestha_Orchard18_slayspik, bellec_slow_2016, huh_gradient_2018, neftci_surrogate_2019, yin_effective_2020, wozniak_deep_2020, zenke_remarkable_2020}.

While translation approaches require the training of a non-spiking proxy network whose connectivity is later translated into a spiking network, the other methods operate directly on \acp{SNN}. 
Variational learning approaches attempt to 
change the distribution of the network output toward a given target distribution by 
minimizing an upper bound on the Kullback-Leibler divergence between the two.
To that end, these models employ stochastic neuron models, typically formulated within the scope of the spike response model (SRM) with escape noise or a stochastic firing threshold  \cite{gerstner_neuronal_2014}.
Variational methods can learn useful representations in hidden neurons and, importantly, the resulting learning rules often have natural interpretations as local learning rules with a global modulatory factor \cite{brea_matching_2013, rezende_stochastic_2014, jang_vowel_2020}. 
However, both low dimensional feedback \cite{werfel_learning_2004} and stochasticity \cite{rezende_stochastic_2014} are known to result in noisy gradient estimates, which can lead to slow convergence and can render learning practically impossible.
    
Surrogate gradient learning avoids such problems by using neuron specific feedback signals as in standard \ac{BP} and dispensing with stochasticity in the forward pass.
Nevertheless, gradients are computed as if noise was present to smooth out the non differentiable binary nonlinearities of spiking neurons.
While a rigorous theoretical formulation of this interpretation still needs to be worked out, it could provide a compelling explanation of why surrogate gradient learning is robust to the choice of nonlinearity \cite{zenke_remarkable_2020}. Within such a framework, different functional shapes of surrogate derivatives may simply correspond to different choices of neuronal noise distributions.

Irrespective of the underlying explanation,
a host of recent studies have established the effectiveness of surrogate gradient learning at scale on various deep \ac{SNN} architectures and diverse tasks and datasets 
\cite{Lee_etal16_traideep, Shrestha_Orchard18_slayspik, yin_effective_2020, neftci_surrogate_2019, bellec_solution_2020, zenke_remarkable_2020, wozniak_deep_2020}.
To that end, surrogate derivatives have been used in combination with variants of both \ac{RTRL} or \ac{BPTT}.

\begin{figure*}[!tbh]
    \centering
    \includegraphics[width = 1.0\textwidth]{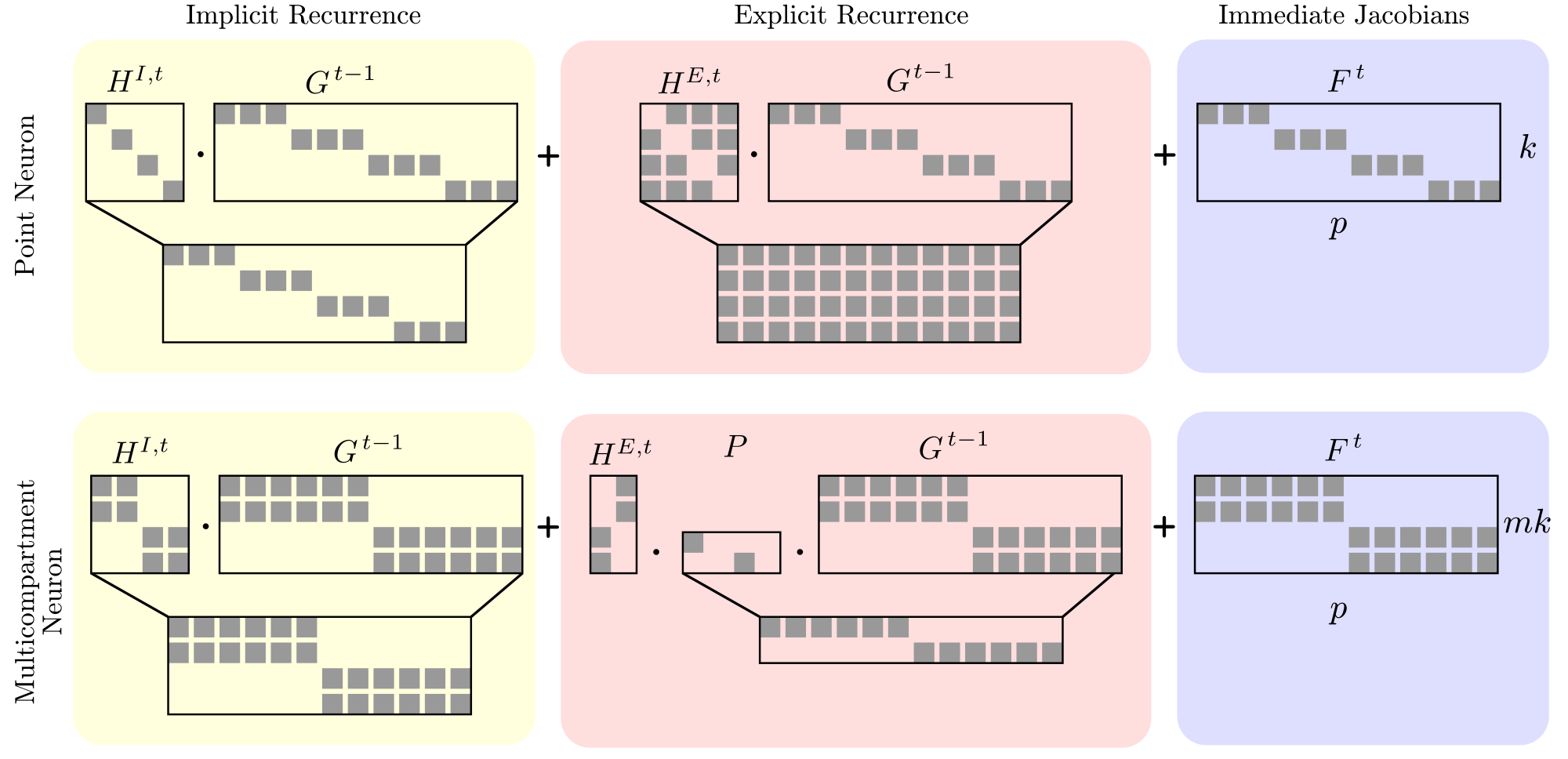}
    \caption{Illustration of the Jacobians involved in the \ac{RTRL} recursion \refeq{eq:rtrl_recursion} for 4 point neurons ($n=3, k=4$, $p=12$, top) and 2 two-compartment neurons $n=3, k=2$, $m=2$, $p=nkm$ (bottom). To distinguish input and output dimensions, the recursion is shown here for the parameters of $W$ only. For parameters $V$, this illustration would look similar, but with $n=k$. (Top) Point Neuron, single compartment neuron. For both cases, it was assumed that $G^{t-1}$ is a step-wise diagonal matrix. This shows the ``mixing'' effect of $H^{\mathrm{E},t}$. By virtue of block matrix algebra, the implicit recurrence preserves the (block) diagonal structure. Hence, $H^{\mathrm{I},t}$ is referred to as implicit recurrence. On the other hand, the off-diagonal terms of $H^{\mathrm{E},t}$ destroy the (block) diagonal property. Here we illustrate $H^{\mathrm{E},t}$ as having no diagonal terms, meaning there are no synaptic self connections.}
    \label{fig:diagonal_part}
\end{figure*}

\subsection{Spiking Neuron Models have Implicit Recurrence}
\label{sec:types_of_recurrence}

To capture the internal dynamics of real neurons, spiking neuron models possess implicit recurrence \cite{neftci_surrogate_2019}.
To understand this concept, we consider
one of the simplest and most widely used spiking neuron models, the \ac{LIF} neuron \cite{gerstner_neuronal_2014}.
\ac{LIF} neurons capture key electrophysiological properties of biological neurons, while being abstract, analytically tractable, and easy to simulate.
The state of a single-compartment \ac{LIF} neuron $i$ is described by its membrane potential $U_i$ which obeys the following temporal dynamics
\begin{equation}
\begin{split}
	\tau_\mathrm{m} \frac{\mathrm{d}U_i}{\mathrm{d}t} = & -(U_i-U_\mathrm{rest}) \\
	&+ R \sum_j W_{ij} S_{j,\mathrm{in}}(t) + R \sum_k V_{ik} S_k(t),
    \label{eq:lif_ode}
\end{split}
\end{equation}
where we have introduced the resting potential $U_\mathrm{rest}$, 
the membrane time constant $\tau_\mathrm{m}$, and the input resistance $R$ \cite{gerstner_neuronal_2014}.
The membrane potential $U_i$ acts as a leaky integrator of the input spike trains $S_{j,\mathrm{in}}(t)=\sum_{k \in \zeta_j} \delta(t^k_j-t)$ where $delta$ is the Dirac delta and the sum runs over all firing times $\zeta_j$ of neuron $j$. 
Output spike trains $S_k(t)$ are defined similarly.
Matrices $W$ and $V$ here define feed-forward and recurrent weight matrices.
On a digital computer, Eq.~\eqref{eq:lif_ode} is commonly integrated using an Euler numerical integration\footnote{Due to the chaotic nature of many \ac{SNN} models, and the dominant discretization error introduced by the simulation time grid, most simulators rely on Euler integration \cite{Goodman_Brette08_briasimu,zenke_limits_2014}.} on a discrete time grid as follows:   
\begin{equation}
    \label{eq:lif_discrete}\\   
    \begin{split}
    U_i^{t+1} = \beta U_i^{t} + (1-\beta) (-U_\mathrm{rest} & + R \sum_j W_{ij} S_{j,\mathrm{in}}^t \\ 
    &+ R \sum_k V_{ik} S_k^t),
    \end{split}
\end{equation}
where we further introduced the decay factor $\beta:=\exp\left(-\frac{\Delta}{\tau_\mathrm{m}}\right)$ with timestep $\Delta$ and the discrete spike trains $S_{j,\mathrm{in}}^t$, $S_j^t$  which are equal to one if the respective neurons $j$ spiked in timestep $t$ and zero otherwise.
Equation~\eqref{eq:lif_discrete} makes the formal connection to our \ac{RNN} example above more obvious.
Similar to a \ac{LSTM} cell, the spiking neuron's leaky membrane potential maintains an internal state through \emph{implicit recurrence} via the decay constant $0<\beta<1$.
The \ac{LIF} (\refeq{eq:lif_discrete}) model is related to the \ac{RNN} and its learning dynamics in \refeq{eq:rtrl},
by replacing $h^t$ with the membrane potential state $U^t$.

To understand the implications of this property for gradient computation and to use it for efficient algorithmic learning implementations, we distinguish between two types of recurrence, namely, explicit and implicit.
In keeping with our convention (cf.\ Eq.\ \eqref{eq:rtrl}), we define the following notation:
\[\label{eq:point_rtrl}
G^{t} = \pd{U^{t}}{\theta},
\,
F^{t} = \pd{U^{t}}{\theta^t},
\,
H^{t} = \underbrace{\pd{U^{t}}{U^{t-1}}}_{H^{\mathrm{I},t}} + \underbrace{\pd{U^{t}}{S^{t-1}}\pd{S^{t-1}}{U^{t-1}}}_{H^{\mathrm{E},t}},
\]
where $H^{\mathrm{I},t}$ and $H^{\mathrm{E},t}$ denote the implicit and explicit recurrent dynamics, respectively. 
This decomposition is motivated by the element-wise nature of the computations inside the network elements.
Implicit recurrence captures the sensitivity to perturbations of the internal neuronal dynamics, such as membrane dynamics.
Explicit recurrence is due to inter-neuronal synaptic connections, like in any vanilla \ac{RNN} model. 
With these definitions, the forward-mode recursion takes the familiar \ac{RTRL} form that exhibits the contributions of the two types of recurrences:
\begin{equation}\label{eq:rtrl_recursion}
\begin{split}
G^{t} &= \underbrace{H^{\mathrm{I},t} G^{t-1}}_\mathrm{Implicit} + \underbrace{H^{\mathrm{E},t} G^{t-1}}_\mathrm{Explicit} + F^{t}\\
\end{split}
\end{equation}
with initial states $G^{0}=0$.

A consequence of element-wise operations within the neuron is that $H^{\mathrm{I},t}$  and $\pd{S^t}{U^t}$
are block sparse Jacobian matrices with a diagonal structure
\reffig{fig:diagonal_part}.

The model descriptions above represent \acp{SNN} in the same form as artificial \acp{RNN}.
However, compared to \acp{RNN}, the timestep used for \ac{SNN} integration is much finer in practice to account for the internal neuronal dynamics. 
Further, a small time step provides a more accurate integration with respect to the continuous-time dynamics and thus the modeled neuromorphic substrate.
Since this timestep is determined by the shortest time scale in the dynamical system, and thus independent of the input data, many \acp{SNN} are trained over hundreds of steps.
This results in as many network instances in memory as there are time steps, which becomes cumbersome for networks of more than a few thousand neurons.
As a result, in recent works, the size of \acp{SNN} trainable by \ac{BPTT} has remained severely limited by the available GPU memory \cite{Shrestha_Orchard18_slayspik}.
Hence, pure reverse-mode \ac{AD}, \emph{i.e} \ac{BPTT} is not suitable to train large \acp{SNN}.
The invariance of \ac{RTRL}'s complexity to the number of time steps~$T$ offers a potential solution to this problem, provided the unfavorable $O(k^3)$ space and $O(k^4)$ time scaling can be mitigated. 
But how can one achieve a reduction in complexity?

Let us, for a moment, ignore the explicit recurrences when computing $G^t$. 
In this case, if $H^\mathrm{I}$ and $G^{t}$ have the same block structure\footnote{$G$ is not square in general, but the same arguments hold for step-wise diagonal matrices shown in \reffig{fig:diagonal_part}.}, the resulting product $H^{\mathrm{I},t}G^{t-1}$ is also block-diagonal.
Thus, only $O(k^2)$ operations are necessary to compute the recursion of the derivative, and $O(k^2)$ memory is required to store the non-zero values of $G^t$. 
Therefore, maintaining the sparseness of the $G^t$ recursion has the potential of reducing the complexity of \ac{RTRL}.
Here, we focus on two solutions to maintain $G^t$ sparse.

The first solution is to work with sparse $H^\mathrm{E}$ matrices, meaning in the case of a single layer \ac{RNN} a sparse matrix $V$ with few non-zero entries.
Sparse connectivity patterns are pervasive in the brain as biological neural networks tend to be locally dense but globally sparse \cite{Ercsey-Ravasz_etal13_prednetw}. %
While a sparse $V$ does not indefinitely maintain the $G^t$ sparse, it does so for a certain number of steps.
This opens up possibilities for sparse $n$-step approximations, allowing to save computation by a factor of the sparsity squared \cite{menick_practical_2020}.
Moreover, sparse connectivity caters to the fact that neuromorphic hardware often relies on sparse connectivity for better memory-efficiency \cite{Pedroni_etal19_memosyna}, which is also mirrored in the hierarchical communication fabric \cite{Moradi_etal17_scalmult, Park_etal17_hieraddr, Davies_etal18_loihneur}. 
Thus, from an implementation standpoint, sparse connectivity matrices are preferable on neuromorphic hardware.
From a performance standpoint, the suitability of sparsely connected networks varies from case to case, and is an intensely studied topic, both in terms of implementation \cite{Gray_etal17_gpukern} and algorithms.
Randomly pruning network weights typically impairs overall network performance unless special care is taken, such as intelligent sparse initialization schemes \cite{frankle_lottery_2018, lee_snip_2019, tanaka_pruning_2020, liu_finding_2020} or dynamic rewiring during the training \cite{bellec_deep_2017, evci_rigging_2019}.
Moving forward, the ``lottery ticket hypothesis,'' which posits the existence of sparse, trainable feed-forward networks without loss in accuracy \cite{frankle_lottery_2018}, is likely to spur further research on sparse \acp{RNN}.

The second solution is to approximate the gradient computation by assuming that certain connections contribute more to the gradient than others. 
For \acp{SNN}, a simple way of doing so is to ignore all contributions of $H^\mathrm{E}$ to the gradient which amounts to assuming that most relevant temporal information is carried forward in time through implicit recurrence, \emph{i.e.}, $H^\mathrm{I}$. 
All the while, the recurrent connections remain in place in the network and contribute to the dynamics. 
But how well do such approximations work?

\begin{figure*}[!tbh]
\begin{center}
\includegraphics[width=1.0\textwidth]{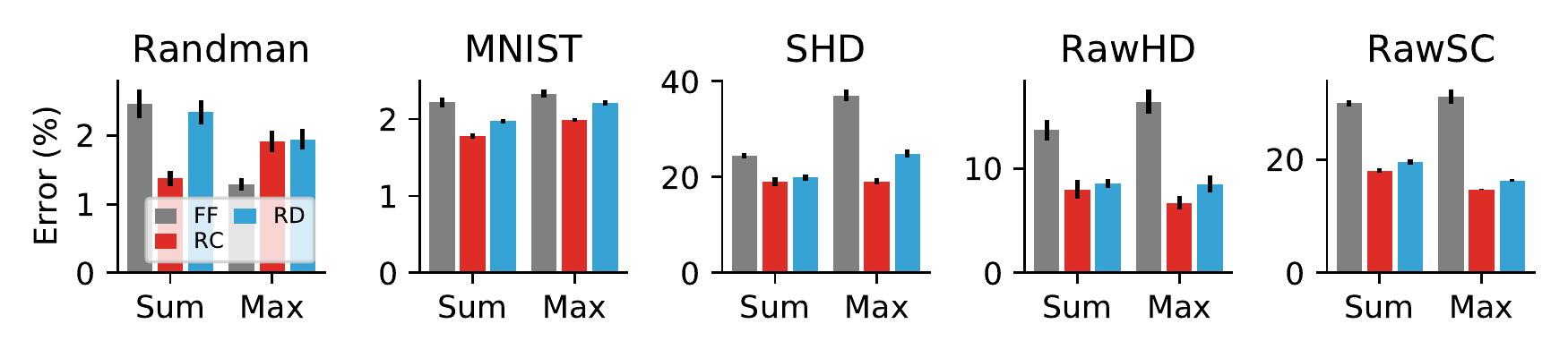}
\end{center}
\caption{Classification error of \acp{SNN} trained on various classification tasks as in \cite{zenke_remarkable_2020}.
Randman: A synthetic spike-timing dependent task based on smooth random manifolds.
MNIST: A time-to-first spike latency encoded version of the MNIST dataset.
SHD: The Spiking Heidelberg Digits dataset \cite{cramer_heidelberg_2019}.
RawHD: The Heidelberg Digits dataset, but using analog current-based input instead of spikes. 
RawSC: Same current-based input, but using the Speech Commands dataset \cite{warden_speech_2018}.
We distinguish between the following types of synaptic connectivity and training approaches. FF (gray): feed-forward \acp{SNN}, RC (red): explicitly recurrent \acp{SNN} trained with surrogate gradients, and RD (blue): ``recurrent detachted'' networks that are architecturally identical to RC, but in which we ignored explicit recurrence for gradient computation (cf.\ $H^\mathrm{E}$).
Finally, we used two different readout configurations to compute the logits for the softmax cross-entropy loss.
We either summed up the readout unit activation over all timesteps (Sum) or computed the Max over all timesteps.
Due to the high computational cost of full \ac{RTRL} and the absence of efficient training libraries, we performed all simulations using \ac{BPTT}
and determined optimal hyperparameters with a grid search comprising more than 8000 simulations. 
For each configuration we selected the ten best models using held-out validation data and computed the mean error and SEM on separate test data.} 
\label{fig:recurrent_detached}
\end{figure*}

\subsection{Implicit recurrence induces approximate, local, and efficient learning rules}

Although ignoring $H^\mathrm{E}$ seems like a drastic simplification, several studies have used it to construct biologically plausible online learning rules as local approximations of \ac{RTRL}.
Empirically, these rules perform well on a number of complex problems either without recurrent connections, like in the case of SuperSpike \cite{zenke_superspike:_2018}, or by ignoring gradient flow through the recurrent synaptic connectivity as done in e-Prop \cite{bellec_solution_2020}, RFLO \cite{murray_local_2019}, and DECOLLE \cite{kaiser_synaptic_2020}. 
These findings suggest that explicit recurrence is either not necessary for many problems, or that ignoring explicit recurrence in gradient computations does not create a major impediment for successful learning, even when such recurrent connections are present.

To disambiguate between these different possibilities, we  
extended previous work \cite{zenke_remarkable_2020} by running additional simulations in which we either ignored or included the contribution of $H^\mathrm{E}$ during \ac{SNN} training.
We then systematically compared the resulting network performance of the two approaches and to networks without any explicit recurrent connections \reffig{fig:recurrent_detached}.
Since these results may be dataset dependent, we repeated this analysis for a range of different classification datasets that required different levels of temporal memory.
The tasks can be coarsely divided according to the duration of their inputs. 
For the Randman and MNIST dataset all input spikes arrive within a short temporal window ($<50\mathrm{ms}$), whereas for the speech processing problems, we considered individual inputs with a duration of $\sim1\mathrm{s}$.

Not surprisingly, the addition of recurrent synaptic connections (RC) to a given \ac{SNN} results in a reduction of error over strictly feed-forward (FF) synaptic connectivity, in most cases \reffig{fig:recurrent_detached}. 
Note, that we did not correct for the larger parameter count of the recurrently connected models.
This observed difference was generally larger for tasks with increased complexity and temporally longer stimuli, consistent with the idea that longer stimuli require longer memory time scales.
However, when the same \acp{SNN} were trained ignoring gradient contributions through explicit recurrence through $H^\mathrm{E}$, error rates increased mildly on most datasets.
Importantly, however, the errors remained lower than for the networks without recurrent synaptic connections.
The effect was largest on tasks that required more temporal memory.

Thus, in the scenarios we tested, the cost of ignoring the contribution of $H^\mathrm{E}$ to gradients is small, all the while the algorithmic benefits are substantial.
As illustrated above, ignoring $H^\mathrm{E}$ yields local online learning rules (cf.\ \eqref{eq:rtrl_recursion}), which are better suited for hardware implementations.
Several online learning approaches therefore make use of this, or similar approximations \cite{murray_local_2018, bellec_solution_2020, kaiser_synaptic_2020}.

The \ac{LIF} neuron model used for the simulations in Fig.~\ref{fig:recurrent_detached} had a two dimensional state. 
One variable was used to model the membrane potential, whereas the other described the time course of exponentially decaying synaptic currents.
However, the scaling properties of the online learning algorithm are not affected by extending it to more complex multi-compartment neuron models, for instance, by incorporating additional slow dynamical variables \cite{bellec_slow_2016, kalchbrenner_efficient_2018, bellec_solution_2020, yin_effective_2020}.

\subsection{Implicit Recurrence of Multi-compartment Neurons can Increase Computational Power}

We now illustrate that \ac{RTRL} applied to multi-compartment neurons results in Jacobians $H^{\mathrm{I},t}$ that are block diagonal and, hence, efficient to train using approximate online algorithms.
This can be formalized by extending the domain of $U^t$ to $\mathbb{R}^{mk}$, where $m$ is the total number of compartments per neuron.
Note, that in this formalism synaptic dynamical variables also count as compartments.
For instance, the ones typically used to implement exponentially decaying conductance or current variables as for the experiments shown in \reffig{fig:recurrent_detached} (cf. ~Appendix~\ref{sec:appendix_rtrl}). 
In a typical multi-compartment model, $S^t \in \mathbb{R}^{k}$ is a function of only one compartment per neuron, which can be written:
\[
S = \Theta(P U^t)
\]
where $P$ is a binary ${k \times mk}$ matrix that selects the spiking compartment.
Without loss of generality, we can assume the first compartment is the spiking one, resulting in the following matrix $P$:
\[
P_{ij} = \begin{cases} 1,\text{ if } j = m\cdot i \\ 0, \text{ otherwise } \end{cases}
\]
Consequently, the dynamics are given by:
\begin{equation}\label{eq:multicomp_rtrl}
\begin{split}
H^{t} &= \underbrace{\pd{U^{t}}{U^{t-1}}}_{H^{\mathrm{I},t}} + \underbrace{\pd{U^{t}}{S^{t-1}}\pd{S^{t-1}}{PU^{t-1}}}_{H^{\mathrm{E},t}}P.\\
\end{split}
\end{equation}
Since only spiking compartments are assumed to be connected with other compartments, the weights $V$ are matrices defined in $\mathbb{R}^{mk\times k}$. Furthermore, we have that $G^t \in \mathbb{R}^{mk \times p}$.
Following the approximate gradient computation where $H^E$ is ignored, $G^{t}$ becomes block diagonal \reffig{fig:diagonal_part}, resulting in $pm$ non-zero entries. 

Thus, adding neuronal complexity by widening the neuronal state space does not change the time complexity of the learning algorithms and does not preclude the use of efficient online learning algorithms.
However, such changes can have dramatic effects on the computational expressivity of the resulting network models and are reflected in the corresponding approximate learning rules \cite{bellec_solution_2020}.
These insights may partially explain why neurobiology uses a diversity of different neuron types with distinct internal dynamics and thus opens up new vistas for exciting future research.

\subsection{Different Modes of Auto-differentiation in Biologically Inspired Learning} 
To train biologically inspired \acp{SNN} and to study the dynamical properties of biological networks discussed above in functional networks, many researchers rely on auto-differentiation frameworks. But because of the shortcomings of \ac{BP} in the context of online learning for neuromorphic applications, increasing attention is given to the approximate online learning methods discussed in the previous sections.
In the following, we give concrete examples of recent work on \ac{SNN} training.
However, it is important to bear in mind that the same algorithms also apply to non-spiking \acp{RNN} when the spiking nonlinearity is replaced with a smooth differentiable function. 

\begin{figure*}[!thb]
    \begin{center}
    \subfloat[Full reverse-mode (BPTT) \label{fig:rnn_unrolled_rev}]{%
    \includegraphics[width=.3\textwidth]{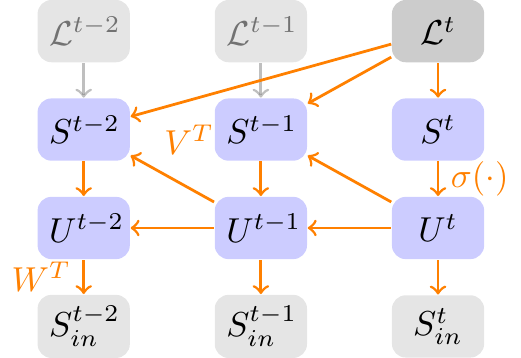}
     }
    \hspace{.25cm}
    \subfloat[Approximate forward-mode (RTRL) \label{fig:rnn_unrolled_for}]{%
    \includegraphics[width=.3\textwidth]{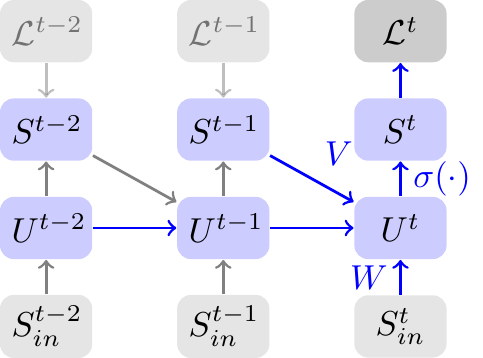}
     }
    \hspace{.25cm}
    \subfloat[Mixed-mode \label{fig:rnn_unrolled_mix}]{%
    \includegraphics[width=.3\textwidth]{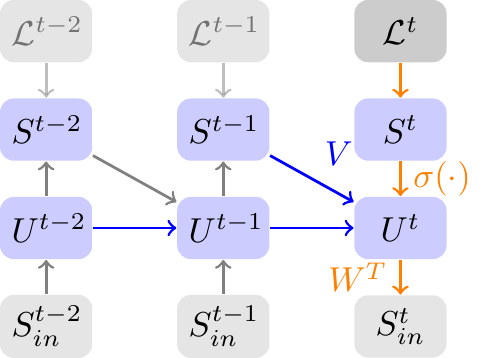}}
    \end{center}
\caption{\label{fig:rnn_unrolled} \textbf{Computational graphs for the gradient using three distinct modes}. Edge colors emphasize gradient computations in reverse (orange) or forward (blue). Grey edges indicate that gradients are not involved. For clarity, only the computations relating to $\mathcal{L}^{t}$ are emphasized.}
\end{figure*}

Specifically, we will show one case of reverse mode \ac{AD} (\ac{BPTT}), one case of forward mode \ac{AD} (\ac{RTRL}), and one example of mixed-mode \ac{AD} \reffig{fig:rnn_unrolled}.
For all three examples, we consider a network consisting of \ac{LIF} neurons which evolve according to the dynamics in \refeq{eq:lif_discrete}.
To simplify the mathematical expressions and without loss of generality, we consider $U_\mathrm{rest}=0$ and $R=1$.
We write the \ac{LIF} dynamics in matrix form as follows:
\begin{equation}
\begin{split}\label{eq:simple_lif_example}
U^{t+1} &= \beta U^t + (1-\beta) (W S_\mathrm{in}^{t} + V S^t),\, S^t = \Theta(U^t).
\end{split}
\end{equation}
Here, we assume $n$ input and $k$ output neurons, and $S_\mathrm{in}^t$ to be the input spike train.

\paragraph{Training with BPTT \reffig{fig:rnn_unrolled_rev}}
We first analyze the case of training this network with \ac{BPTT}. 
The hidden state is $U^t$ and the dynamics and immediate Jacobian takes the form: 
\[
\begin{split}
    H^t &:= \pd{U^t}{U^{t-1}} = \beta + (1-\beta)V \sigma'(U^{t-1})\\
    D^t &:= \pd{\mathcal{L}^t}{U^t} = \sigma'(U^t)
\end{split}
\]
where we have assumed the smooth surrogate function $\sigma$.
Using the \ac{BPTT} recursive expression for $C^t$ (which includes dynamics $H^t$, \refeq{eq:bptt}), the gradients for $W$ and $V$ become:
\begin{equation}
\pd{\mathcal{L}^t}{W}  =  C^t S_\mathrm{in}^{t},\,  \pd{\mathcal{L}^t}{V}  = C^t S^{t}. \\
\end{equation}
Thanks to the native support of reverse mode \ac{AD} in machine learning frameworks, this spiking neuron model can be implemented and differentiated like \acp{RNN}.
This approach has been applied successfully to train general \acp{SNN} models \cite{Shrestha_Orchard18_slayspik, bellec_long_2018, cramer_heidelberg_2019, yin_effective_2020, huh_gradient_2018, Thiele_etal19_spikann-, Lee_etal20_enabspik}.

\ac{BPTT} has the advantages that it does not restrict the dynamics, connectivity patterns, and loss function.
However, these advantages come at the cost of a large memory footprint and temporal non-locality.
Thus, for applications using small networks of some thousands of neurons that do not require online learning, \ac{BPTT} is the method of choice.
For larger networks, the memory overhead becomes impractical, and other modes of gradient computation described below may be necessary.

\paragraph{Training with Sparse \ac{RTRL} \reffig{fig:rnn_unrolled_for}}
As discussed above, it is possible to reduce the complexity of \ac{RTRL} by keeping $G^t$ sparse. 
We demonstrate this in the case of the \ac{LIF} neuron.
Using the \ac{RTRL} recursion (\refeq{eq:rtrl}), the gradient of $W$ obeys:
\begin{equation}
    \begin{split}
        G_W^t =  ( \beta + \underbrace{(1-\beta)V \sigma'(U^{t-1})}_\text{Explicit rec.}) G^{t-1}_W + F^t.
    \end{split}
\end{equation}
If we neglect explicit recurrence in the above expression, all involved matrices remain block sparse as shown in \reffig{fig:diagonal_part}. 
The recursion can be compactly written in vector form by defining the following $k$-vector trace:
\begin{equation}
\begin{split}\label{eq:mixed_simple}
  Q^{t+1}_\mathrm{in} &= \beta Q^{t}_\mathrm{in} + S^{t}_\mathrm{in}, \\
\end{split}
\end{equation}
where we used a different variable $Q_\mathrm{in}$ to emphasize that it represents a vector, rather than a tensor.
Since the $Q_\mathrm{in} \in \mathbb{R}^{n}$ terms are dense $n$-vectors, for gradients with respect to $W$, the \ac{RTRL} recursion is simplified from $O(np)$ memory to $O(n)$ memory. 
Gradients are then computed in the familiar three-factor form:
\begin{equation}\label{eq:3frule}
\pd{\mathcal{L}^t}{W}  = \pd{\mathcal{L}^t}{S^t}  \sigma'(U^t) ~ Q_{in}^{t},
\end{equation} 
and similarly for $V$ gradients: 
\[ 
\pd{\mathcal{L}^t}{V} = \pd{\mathcal{L}^t}{S^t}  \sigma'(U^t) ~ Q^{t},
\]
where $Q^t \in \mathbb{R}^{k}$ is defined in analogy to $Q_\mathrm{in}^t$, but replacing $S_\mathrm{in}^t$ with $S^t$.
Note the above factorization to the dense $k$-vector form is only valid for linear \ac{LIF} neurons with instantaneous approximations of the loss function \cite{kaiser_synaptic_2020}. 
However, in the case of nonlinear neuronal dynamics, such as spike-triggered adaptation or for certain loss functions (see Section \ref{sec:losses}), $O(k^2)$ traces may be necessary, leading to quadratic scaling with the number of neurons \cite{zenke_superspike:_2018, bellec_solution_2020}.

\paragraph{Mixed-mode}
It is possible to combine elements of both \ac{BPTT} and \ac{RTRL} for training \acp{RNN} and \acp{SNN}.  
We call this situation mixed-mode \ac{AD} \reffig{fig:rnn_unrolled_mix}.
The portion of the graph from $S^t_\mathrm{in}$ to $\mathcal{L}^t$ in \reffig{fig:rnn_unrolled} may involve several steps. 
Such cases can occur for example in convolutional networks with pooling layers, linear readout layers and multiple recurrent layers \cite{Bohnstingl_etal20_onlispat}.
In \cite{kaiser_synaptic_2020}, for example, loss functions were defined based on random combinations of max-pooled outputs of a convolutional layer consisting of spiking neurons.
If these steps are instantaneous, \emph{i.e.}, they do not explicitly depend on past states, the immediate Jacobians $F^{t}_W$ and $F^{t}_V$ can be computed online using \ac{BP} within a single timestep.
All other gradients can be accumulated over time by virtue of the \ac{RTRL} recursion.
Up to rounding errors, the numerical result will be exactly the same as sparse \ac{RTRL} discussed earlier, but the memory footprint of the implementation differs and may be more favorable. 
The recent \ac{DCLL} \cite{kaiser_synaptic_2020} is one example of a mixed-mode \ac{AD} implementation using spatially and temporally local loss functions. 
It combines forward mode \ac{AD}, to achieve temporal credit assignment, with reverse mode \ac{AD} for spatial credit assignment. 
This allows the convenient use of existing autodifferentiation tools, while combining it with the more favorable scaling properties of \ac{BP} in space.  
Bohnstingl \emph{et al.} \cite{Bohnstingl_etal20_onlispat} lay the ground to extend this idea to multiple recurrent layers, and find that estimating gradients for layer $l$ using solely the recurrence relation of that layer, and ignoring others results in good learning performance and low complexity.

\section{Loss functions for online learning}
\label{sec:losses}

So far, we have focused on learning algorithms that permit efficiently computing loss gradients in an online manner to learn from temporal data. 
The applicability of these algorithms, however, hinges on the ability to also compute the loss functions in an online manner, which is not always possible.
In the following, we discuss the desiderata of loss functions for online real-time learning and point out directions where future research is required.

\begin{figure*}[tbp]
    \centering
    \includegraphics[width=1.0\textwidth]{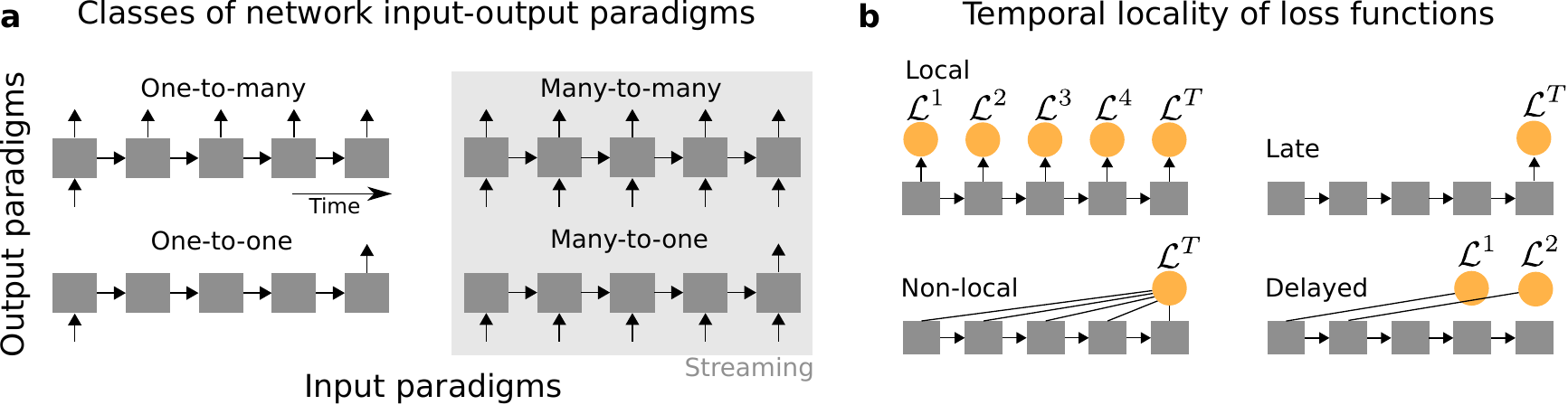}
    \caption{\textbf{a}~Schematic view of different computational graphs of \acp{RNN} illustrating the main different input-output paradigms that typically exist in \acp{RNN}. For streaming data purposes, the many-to-one and many-to-many mapping in the gray shaded box are most relevant.
    \textbf{b}~Illustration of different computational graphs used for computing loss functions. Online learning algorithms such as \ac{RTRL} require temporally localized loss functions that can be evaluated at every timestep, which trivially generalizes to the regime in which many local losses are zero, e.g., $\mathcal{L}^t$ for $t<T$.
    However, non-local loss functions such as the sum or the maximum over network outputs, as well as, delayed arrival of labels that cause the loss evaluation to be delayed pose challenges for online learning.
    }
    \label{fig:loss_functions}
\end{figure*}

Since loss functions are defined at the output of a network, let us briefly review what input-output paradigms exist for \acp{RNN}. 
We can coarsely divide neural network processing by specifying whether they require an input at every timestep or only once. Similarly, we can distinguish between networks that yield one output at the end of processing as opposed to networks that produce an output in every timestep.
Networks that receive a single input at the beginning produce a single output at the end, are very similar to feed-forward neural networks in that they provide a one-to-one mapping (Fig.~\ref{fig:loss_functions}a). Similarly, we can construct networks that output a trajectory in response to a single command input, which corresponds to a one-to-many mapping.
For streaming data, networks must be able to consume a temporal sequence of inputs. Thus, these networks have \emph{many} inputs and 
can be further separated into many-to-many or many-to-one mappings.

For our above derivations pertaining to \ac{RTRL}, we assumed a gradient of a loss function $\mathcal{L}$ defined as a sum over time $\mathcal{L}=\sum_t\mathcal{L}^t$ of temporally localized loss functions $\mathcal{L}^t$. 
We assumed further that these localized loss functions can be computed immediately after the corresponding network timestep is computed, \emph{i.e.} online. 
It is easy to see how such local functions can be defined on the output of a many-to-many network (Fig.~\ref{fig:loss_functions}b).
Similarly, this also covers the many-to-one case since we can simply set all losses for $t<T$ to zero ($\mathcal{L}^t=0$) and retain only one \emph{late} value $\mathcal{L}^T$.

In the context of \acp{SNN}, several studies have focused on the many-to-many setting in which the network has to learn a predefined output trajectory. 
A classic example is FORCE learning 
which applies the recursive least squares algorithm which can be done forward in time and online \citep{sussillo_generating_2009, nicola_supervised_2017}.
A related approach is FOLLOW learning \citep{gilra_predicting_2017} which is another online learning approach for \acp{SNN} that relies on local learning rules to learn a forward model of arbitrary dynamical systems.
However, not all learning tasks require learning an output trajectory and hence temporal locality of the loss function is not the rule, but rather the exception. 
To solve classification problems, for example, one often considers a non-local loss that takes into account extended periods of network activity (Fig.~\ref{fig:loss_functions}b).
For instance, in seminal work G\"utig \emph{et al.} \citep{gutig_tempotron:_2006, gutig_spiking_2016} trained a spiking neuron, the Tempotron, as a binary classifier on input spike trains with a sparse temporal code. 
Crucially, the Tempotron can freely decide when to spike within a given temporal window.
This has the advantage that targets or labels do not have to be given to a learning system with temporal acuity.
To achieve this, the authors introduced a loss that depends on the maximum of the neural membrane potential during an entire trial. 
Hence, the loss function can only be evaluated once the trial is completed. 

A similar approach was later extended to train multi-layer \acp{SNN} by either computing the maximum over time or the sum over time of dedicated readout units \citep{cramer_heidelberg_2019,zenke_remarkable_2020} or by simply summing over the number of output spikes \citep{Shrestha_Orchard18_slayspik, bellec_long_2018, Stewart_etal20_onlifew-}. 
Although this approach can learn powerful classification models, non-local loss functions are not suitable for online learning since their gradient can only be evaluated after a trial is completed.
Although similar to the late loss case, they are different in that their value depends explicitly all timesteps (Fig.~\ref{fig:loss_functions}b).

This temporal non-locality has far-reaching consequences for all classification and pattern detection tasks, because
it means that all network activity has to be stored until the loss function can be evaluated. 
Thus gradient computation is \emph{locked} until the last timestep of a given trial is evaluated. 
A similar situation arises when targets or labels are delayed because this also affects the corresponding loss function evaluations (Fig.~\ref{fig:loss_functions}b). 
While locking is the norm in the context of \ac{BPTT}, 
online algorithms such as \ac{RTRL} may lose their real-time character due to it.
Therefore loss functions that can be evaluated online are crucial to successful online learning. 
Although research on this issue is still limited, one possible remedy to avoiding delays and locking is to fold the non-locality back into the network and to rely on a loss function that can be written as a sum of temporally localized losses.

One example builds on the van Rossum distance which was developed as a distance metric for spike trains \citep{van_rossum_novel_2001}. 
As a spike train distance metric, it was recognized early on as a suitable loss function for training \ac{SNN} to produce precisely timed output spikes \citep{gardner_supervised_2016}, but as we will see in a moment, it is not limited to precisely timed output spike trains.
For a given spike train $S(t)=\sum_k \delta(t_k-t)$ and an associated target spike train $S^*(t)$, the van Rossum distance is given defined as 
\begin{equation}
\mathcal{L} = \frac{1}{2} \int_{-\infty}^\infty \mathrm{d}t \left( \epsilon \ast S(t) - \epsilon \ast S^*(t) \right)^2,
\label{eq:van_rossum_distance}
\end{equation}
where $\ast$ denotes the temporal convolution of the spike trains with the kernel function $\epsilon$.
The trick is now to define a filtered spike train $Y=\epsilon \ast S(t)$ as the network output and similarly a target network $Y^*=\epsilon \ast S^*(t)$.
The kernel $\epsilon$ can be arbitrary, but choosing a causal kernel is critical for all online implementations. 
In practice, we choose a kernel that can be easily implemented as a simple dynamical system \citep{gardner_supervised_2016,zenke_superspike:_2018} 
that allows the online evaluation of the term in parenthesis (cf. Eq.~\eqref{eq:van_rossum_distance}).
Canonical choices are exponential or double exponential functions that are straightforward to implement with one or two additional ordinary differential equations for each output, e.g, $\tau\pd{z}{t}=-z+S(t)$ with the kernel time scale $\tau$.
These manipulations allow us to write 
$\mathcal{L} = \frac{1}{2} \int_{-\infty}^\infty \mathrm{d}t \left( Y(t) - Y^*(t) \right)^2$.
Finally, by further reducing the integral to a sum over discrete timesteps, this expression can be expressed as
$\mathcal{L}=  \frac{1}{2} \sum_t \left( Y^t - Y^{*,t} \right)^2 = \sum_t \mathcal{L}^t$, a sum of local losses.
Finally, it is not necessary to provide a target spike train. Instead, it suffices to simply provide a target time series $Y^{*,t}$ which can be interpreted as an instantaneous target firing rate.

As we already alluded to earlier, the van Rossum distance does not necessarily learn precisely time spikes. Rather, the extent to which the van Rossum distance punishes temporal misalignment can be smoothly adjusted by the time horizon of the $\epsilon$ kernel.
For instance, if one chooses a small timescale $\tau$ for the exponential kernel function $\epsilon$, this allows for learning precisely timed output spike trains. 
However, a large choice of $\tau$ will result in a reduction of temporal spike alignment of individual spikes with their targets. The distance then smoothly interpolates between spike-timing code and rate codes. 
Another interesting property of the van Rossum distance, in particular when used with causal kernels, is that causal kernels induce an effective delay, which acts as an eligibility trace \citep{ zenke_superspike:_2018}. Eligibility traces are found in neurobiology \cite{gerstner_eligibility_2018} and importantly they solve the distal reward problem by bridging the delay between a network output and the error feedback that arrives later in time \cite{izhikevich_solving_2007}.

Thus, at the expense of temporal precision, the van Rossum distance can introduce temporal memory and accommodate label delay during learning \citep{zenke_superspike:_2018} while at the same time allowing to compute temporally localized losses.  
Finally, computing simple kernels such as exponential or double exponential for each network output corresponds to adding additional filtering layers to the network.
This layer implies another set of diagonal Jacobians for which we have already seen how their gradients can be computed using \ac{RTRL}.

While the van Rossum distance solves some challenges in the streaming data setting in which labels can only be assigned coarsely in time, a number of important issues remain open.  
Here future work, possibly building on aggregate label losses \cite{gutig_spiking_2016, pan_efficient_2019} or \ac{CTC} losses \citep{Graves12_supesequ}, provided these approaches can be made online-capable, may offer possible solutions.

In summary, online gradient algorithms based on \ac{RTRL} require temporally localized losses.
While such losses are the standard for learning output trajectories, there are situations in which temporally non-local loss functions are more natural choices (e.g., classification problems).
Fortunately, a conversion to online-enabled losses is possible for some non-local loss functions, as we illustrated in the example of the van Rossum distance.
But future research efforts are required to firmly establish online-capable loss functions for situations in which labels are only loosely aligned with streaming input.

\section{Implementation Strategies in Neuromorphic Hardware}
How can the gradient-based learning strategies discussed in this article guide the development of a neuromorphic learning substrate? 
Memory is generally a limiting factor in neuromorphic hardware. 
In \acp{SNN}, the need for online learning in combination with such capacity limits makes forward mode accumulation particularly compelling. 
Sparse \ac{RTRL} equates to one trace per parameter, which can be realized in hardware by replicating the circuits and storage for weight updates. 
For instance, for each connection in Intel Loihi Research Chip, up to two states that evolve as functions of the post- and pre-synaptic states are possible \cite{Davies_etal18_loihneur}.
These states and their dynamics are implemented similarly to the weight update dynamics.
Intel Loihi is thus in principle compatible with a sparse forward-mode \ac{AD} learning scheme.
However, several significant open challenges remain for its implementation.
First, unlike weight (parameter) updates which can be carried out in an event-based fashion, traces are dynamical states per connection that must be updated at every time step.
Because scale in neuromorphic hardware and software simulations is often achieved by storing the synaptic traces at the neuronal level \cite{Brette_etal07_simunetw}, computing traces per connection would significantly increase the computational burden.
Second, the memory overhead becomes significant for large networks with shared weights, for instance in convolutional architectures, because weight sharing does not imply trace sharing.
Although the idea of weight sharing is counter intuitive to a neuromorphic implementation, digital neuromorphic hardware can take advantage of it \cite{Davies_etal18_loihneur,Merolla_etal14_millspik}.

One solution to the above problems is to use the sparse approximation described above (cf.\ \refeq{eq:mixed_simple}), previously implemented as part of DECOLLE \cite{kaiser_synaptic_2020} and subsequently for ``LIF neurons'' in \cite{bellec_solution_2020}, resulting to one trace per axon.
In this approximation, synaptic traces use the same state as the synaptic currents for learning.
This implies only constant ($O(1)$) memory overhead for learning, which significantly simplifies the neuromorphic hardware implementation. 
\cite{Payvand_etal20_errothre} describe a DECOLLE crossbar implementation that leverages the sharing of the learning and inference signals, while eliciting updates in a temporally sparse, error-driven fashion.
Furthermore, the learning dynamics are potentially immune to mismatch in the synaptic dynamics, since the same signal is used for computing the forward pass and gradient dynamics.
The gradient-based learning of \acp{SNN} induces a three factor rule (\refeq{eq:3frule}), comprising one term to compute the loss gradient ($\pd{\mathcal{L}}{S^t}$) and two terms for the network states ($\pd{S^t}{\theta}$).
\cite{Payvand_etal20_errothre} exploits this factorization in a neuromorphic design comprising two types of cores: processing cores and neuromorphic cores. 
Processing cores are general-purpose processors that compute the loss gradients, and neuromorphic cores compute the network states and their gradients. 
A similar strategy was demonstrated on the Intel Loihi using the on-chip three Lakemont x86 cores \cite{Stewart_etal20_onlifew-}.
This separation imparts significant flexibility to the hardware, as loss functions are often task dependent but network architectures tend to be generic.

Despite the advantages of online algorithms, reverse mode \ac{AD} remains an important reference and a tool for training \acp{SNN} off-line and off-chip. 
One strategy for digital neuromorphic chips is to use a functional simulator of the dynamics \cite{Esser_etal16_convnetw}, and train it using conventional deep learning techniques (GPUs and \ac{BPTT}).
Functional simulators and \ac{BPTT} were also used to pre-train networks for subsequent learning on-chip \cite{Stewart_etal20_onlifew-}.
Due to device-to-device variation, functional simulators of mixed signal hardware require calibration, for example by system identification \cite{Neftci_etal11_systmeth,Bruederle_etal11_compwork}. 
This calibration scheme is costly and often imprecise, because the limited access to the chips' internal states dictates several approximations.

Hardware-in-the-loop approaches can partially overcome this limitation by using the hardware substrate to compute the forward pass and computing synaptic updates in software. 
Recent work demonstrated successful learning on an accelerated neuromorphic VLSI substrate using a chip-in-the-loop approach \cite{Cramer_etal20_traispik}, as well as in phase-change memory \cite{wozniak_deep_2020}.
While this approach can self-correct for any remaining device mismatch, it requires dedicated on-chip circuitry, e.g., sampling ADCs, to read out the internal voltages.
Additionally, the external computation of updates poses a significant performance bottleneck which renders this strategy often too slow for real-time or accelerated learning. 
One solution that has not been fully explored is to pre-train networks on an approximate functional simulator of a mixed signal chip, and fine-tune on chip, as in \cite{Stewart_etal20_onlifew-}.
Regardless of which training strategy is employed, the methods based on reverse mode \ac{AD} are generally limited by memory.
Hence, mixed mode \ac{AD} and other advanced \ac{AD} methods will presumably play a central role in reducing the memory footprint and thereby improving the performance and applicability of \acp{SNN} training to dedicated neuromorphic substrates.

\section{Discussion}

This article has reviewed common bottlenecks encountered when applying gradient-based learning to neuromorphic architectures that require online computation of gradients. 
Using the example of \acp{SNN}, we have shown that many recently proposed learning algorithms for online learning are approximate variants of \ac{RTRL}. In its exact form, \ac{RTRL} is an online algorithm to compute gradients in \acp{RNN}, but it is computationally expensive.  
However, when used in combination with temporally local losses and biologically inspired neural architectures such as \ac{LIF} neurons, it is possible to find effective approximations that reduce \ac{RTRL}'s computational cost substantially while retaining good learning performance.  
We have shown that such approximations are inspired by biological neurons whose implicit recurrence structure induces block sparse or approximately block sparse Jacobians, allowing to speed up gradient computation while simultaneously reducing the memory footprint. These conceptual links expose a clear path forward toward building more efficient online learning algorithms for neuromorphic devices.

We further elaborated on the relationship between gradient-based learning in \acp{SNN} and the different modes of \acl{AD}. 
Due to sufficient memory and the lower computational cost on von Neumann computers, deep learning has primarily focused on reverse mode accumulation or \ac{BPTT}.  However, when combining appropriate architectures with approximate \ac{RTRL}, we expect a renaissance of forward-mode \ac{AD} to empower non-von Neumann computers and streaming applications requiring online learning. Mixed-mode \ac{AD} is a particularly exciting direction for implementing learning in biological neural networks as it efficiently reduces complexity by exploiting the Jacobians' sparseness. This idea resonates with a widespread trend in deep learning accelerators: To trade-off compute against memory, as evidenced by advanced \ac{AD} techniques on manycore processors the Cerebras CS-1 and the Graphcore IPU.  

\subsection{Spatial and Temporal Scales in Gradient-Based Learning}
Computation is a physical process that extends across multiple spatial and temporal scales. Practical learning algorithms have to take this multi-scale behavior into account. 
The finite truncation length in \ac{BPTT} defines the time span (memory) over which the learning algorithm can efficiently navigate between timescales, even when the network itself supports such dynamics (\emph{e.g.}, through working memory or gating mechanisms in \acp{LSTM} \cite{oreilly_making_2006}).
Thus, prematurely truncated gradients can be harmful.
Memory and stability go hand-in-hand in dynamical systems, including \acp{RNN}. 
Miller and Hardt \citep{Miller_Hardt19_stabrecu} argue that \acp{RNN} are trained to operate in the stable regime in which gradients vanish for stable learning.
However, stability can often be at odds with long-term memory \citep{hochreiter_long_1997}.
Therefore, many working memory models in neuroscience rely on multistability and attractor states to implement long-term memory \cite{Amit92_modebrai, litwin-kumar_formation_2014, zenke_diverse_2015}, with regions of state space in which gradients either vanish \citep{bengio_learning_1994} or become very large \cite{Rutishauser_etal11_collstab}.
\ac{LSTM} networks explicitly overcome this shortcoming by including long timescales in their architecture thereby avoiding vanishing gradients \citep{hochreiter_long_1997}.
This twist, however, requires training procedures that can similarly bridge such long time horizons, which is where truncated \ac{BPTT} can easily reach its limits.
An interesting question is whether training with approximate forms of \ac{RTRL} can offer advantages in training networks with such \emph{longer} short-term memory.
Thus, if remaining issues related to the bias of approximate variants of \ac{RTRL} and stability can be addressed at scale, these findings will open new vistas in computing with non-von Neumann computers.

\subsection{Solutions to the Weight Transport Problem}
Another issue that plagues neuromorphic implementations of 
\ac{BP} is that reverse accumulation requires access to the transposed weight matrices~$W$ and~$V$. 
This requirement has also been referred to as the weight transport problem \cite{grossberg_adaptive_1987}.
The weight transport problem implies a bidirectional flow of information, which is biologically implausible, and, crucially, difficult to realize in any physical system \cite{baldi_learning_2018, lillicrap_random_2016}.
The problem is that to compute the gradient, error information needs to flow backward through the same connections that are used in the forward pass.
This creates a major impediment for any physical system, be it biological or neuromorphic, in which information flow is directed. 
Overcoming this limitation often requires an explicit learning channel that implements these backward weights or weight transposes \cite{lillicrap_random_2016, baldi_learning_2018} and ideally keeps them synchronized across nodes or different physical locations of a physical network implementation. 
A number of studies have shown that such synchronization can be achieved, for instance, by learning the backward weights \cite{kolen_backpropagation_1994, akrout_deep_2019, amit_deep_2019}.

In any case, the mere existence of such backward connections for spatial credit assignment still has a real physical price as they require space on the chip and consume energy, thus raising the question of whether one could dispense with them entirely.
Interestingly, \ac{RTRL} and its approximate variants do suggest ways forward toward viable alternatives. 
Since \ac{RTRL} preserves causality with respect to the forward dynamics, the temporal weight transport problem does not apply. 
Unfortunately, exact \ac{RTRL} still requires synapses to know the state of all other neurons and synapses in the network, thus still posing a weight transport problem.
The block sparse Jacobian approximations we discussed in this article are not straight forward to extend to learning across multiple layers.
Nevertheless, the inherently causal nature of \ac{RTRL} and the recent success of local loss functions for training \acp{SNN} \cite{nokland_training_2019,  Mostafa_etal18_deepsupe, kaiser_synaptic_2020, Bohnstingl_etal20_onlispat} provide exciting avenues of future research to spatially assign credit and circumvent the weight transport problem in multilayer networks.

\medskip
While a plethora of neuromorphic platforms, which mimic the brain's computational substrate, have matured over the years, seeing these solutions thrive in real-world applications will require them to learn.
To tackle this challenge, we should turn once more to biology. Taking inspiration from the brain will allow us to develop neuromorphic learning algorithms toward tomorrow's neuromorphic computers.

\section*{Acknowledgments}
This work was supported by the National Science Foundation under grant 1652159 and 1823366 (EN) and the Novartis Research Foundation (FZ). %

\small
\bibliographystyle{IEEEtran}
\bibliography{fzenke,biblio_unique_alt}

\appendices

\section{Derivation of local learning approximations for \acp{SNN} from RTRL}
\label{sec:appendix_rtrl}

\ac{RTRL} is a special case of forward mode \ac{AD} applied to \acp{RNN}. For a population for spiking \ac{LIF} neurons with current based exponential synapses the discrete time dynamics are given by:
\begin{equation}
\begin{split}
	U_i^{(l)}[t+1] & = \beta U_i^{(l)}[t] + I_i^{l}[t]\\
	I_i^{(l)}[t+1] & = \sum_j W_{ij}^{(l)} S_j^{(l-1)}[t] +  \sum_j V_{ij} S_j^{(l)}[t]
    \label{eq:mem_discrete_time}
\end{split}
\end{equation}
The output of neuron $U_i^{(l)}$ is given as the spike train $S_i^{(l)}=\Theta(U_i^{(l)})$.
To compute gradients with respect to this spike train output, we first define
\begin{equation}\label{eq:definition_P}
	P_{ijk}^{(l,m)}[t]     = \frac{\partial} {\partial W_{ij}^{(m)}}  U_k^{(l)}[t]\\
\end{equation}
and take the derivative  global loss function $\mathcal{L}$ with respect to the parameters:
\begin{equation}
\begin{split} \label{eq:forward_mode_differentiation}
  \Delta {W_{ij}^{(m)}} &\propto \Dp[{\mathcal{L}[t]}]{W_{ij}^{(m)}} \\  \Dp[{\mathcal{L}[t]}]{W_{ij}^{(m)}} & = \sum_k
	\Dp[{\mathcal{L}[t]}]{S_k^{(L)}[t]} \sigma'(U^{(l)}_k[t+1]) P_{ijk}^{(L,m)}[t]\\
	P_{ijk}^{(l,m)}[t+1] &=  \left( \beta P_{ijk}^{(l,m)}[t] + \frac{\partial}{\partial W_{ij}^{(m)}}  I_{k}^{(l)}[t] \right)\\
	\frac{\partial} {\partial W_{ij}^{(m)}}  I_{k}^{(l)}[t+1] & =
	\sum_{j'} V_{ij'}^{(l)}	\sigma'(U^{(l)}_{j'}[t]) P_{ijj'}^{(l,m)}[t]\\& \, +
	\underbrace{\sum_{j'} W_{ij'}^{(l)} \sigma'(I^{(l)}_{j'}[t]) P_{ijj'}^{(l-1,m)}[t]}_\mathrm{explicit} \\ 
	& + \delta_{lm} S_j^{(l-1)}[t]. \\
\end{split}
\end{equation}
Similar equations can be obtained for $\Delta {V_{ij}^{(m)}} \propto \Dp[{\mathcal{L}[t]}]{V_{ij}^{(m)}}$.
The terms underwritten with ``explicit'' introduce non-locality to the learning that is difficult to compute, since it depends on the history of all other neuron in the network. 
If these terms are dropped, the indices $i$ and $k$ from $P$ become unnecessary since no term on the right-hand side of depends on those indices.
Thus, for purposes of gradient computation, the indirect influence of all these interactions is ignored, whereas interactions through the implicit recurrence, the memory within the neuron is maintained.
The price of performing such an approximation on task performance is often surprisingly low and it seems that neural networks can nevertheless take advantage of their explicit recurrent connectivity (cf.\ Fig.~\ref{fig:recurrent_detached}).

\end{document}